\documentclass[10pt, letterpaper, journal, twoside]{IEEEtran}
\IEEEoverridecommandlockouts    
\usepackage{algorithm} 
\usepackage[margin=0.679in, top=0.69in]{geometry}

\usepackage{graphicx}
\usepackage{tikz}
\usepackage{pgfplots}
\usepackage{pgfplotstable}
\usepgfplotslibrary{groupplots}

\usepackage{graphics} 
\usepackage{cite}
\usepackage[T1]{fontenc}
\usepackage{amsmath, bm} 
\usepackage{amssymb}  
\usepackage{hyperref}
\hypersetup{
    colorlinks=true,
    linkcolor=blue,
    filecolor=magenta,      
    urlcolor=blue,
}

\usepackage{float}
\usepackage{cancel}

\usepackage{numprint}
\usepackage{comment}

\usepackage{color, colortbl}
\definecolor{LightCyan}{rgb}{0.88,1,1}
\usepackage{multirow}
\usepackage{siunitx}
\usepackage{array}
\DeclareSIUnit\cell{cell}
\DeclareSIUnit\cells{cells}
\DeclareSIUnit\trees{trees}
\usepackage{amsmath,mathtools}
\usepackage[nodisplayskipstretch]{setspace}
\usepackage{booktabs}
\usepackage{hyperref}
\usepackage{xcolor}
\usepackage{multirow}
\usepackage{enumitem}
\usepackage{subcaption}
\usepackage{caption}

\setlist[itemize]{nosep, leftmargin=*}
\setlist[enumerate]{nosep, leftmargin=*}
\setlist{nosep}
\allowdisplaybreaks

\usepackage{microtype}
\makeatletter

\def\@titleheader{}

\newcommand{\titleheader}[1]{\gdef\@titleheader{#1}}

\AtBeginDocument{%
  \let\oldtitle\@title
  \gdef\@title{%
    \begin{center}
    \normalfont\large
    \@titleheader
    \end{center}
    \vskip1em
    \oldtitle
  }
}

\makeatother

\usepackage{titlesec}

\titleformat{\subsubsection}[runin]{\itshape}{\arabic{subsubsection})}{0.5em}{}

\titlespacing*{\subsubsection}{\parindent}{0pt}{*1}
\titlespacing*{\subsection}{0pt}{\baselineskip}{0pt}
\titlespacing*{\section}{0pt}{\baselineskip}{0pt}

\titlespacing*{\section}{0pt}{*1}{*1}
\titlespacing{\subsection}{0pt}{*1}{*1}

\usepackage{enumitem} 
\usepackage{wrapfig}

\usepackage{tikz}
\usetikzlibrary{shapes,arrows}
\usepackage{verbatim}
\usetikzlibrary{positioning}

\usetikzlibrary{external}
\usepackage[normalem]{ulem} 
\usepackage{tabstackengine}
\usepackage{algorithm}
\usepackage[end]{algpseudocode} 
\usepackage{etoolbox}
\makeatletter
\patchcmd{\@maketitle}
  {\centering}
  {\centering
   \vspace*{-2em}
   {\normalfont\footnotesize
   \textcolor{gray}{This paper has been accepted for publication at the 2026 IEEE International Conference on Robotics and Automation}
   \textcolor{blue}{\href{https://2026.ieee-icra.org/}{(ICRA 2026)}}\par}
   \vspace{1.2em}
  }
  {}{}
\makeatother
\usepackage[font=small, skip=2pt]{caption}
\usepackage[skip=3pt]{subcaption} 

\usepackage{pgfplots}
\usepackage{grffile}
\pgfplotsset{compat=newest}
\usetikzlibrary{plotmarks}
\usetikzlibrary{arrows.meta}
\usepgfplotslibrary{patchplots}
\usepackage{amsmath}

\usepackage{bbding} 
\usepackage{colortbl}
\usepackage{hhline}
\usepackage{makecell}
\usepackage{booktabs}

\begin{document}

\title{\LARGE \bf \vspace*{0.15in} LPV-MPC for Lateral Control in Full-Scale Autonomous Racing}

\author{Hassan Jardali, Ihab S. Mohamed, Durgakant Pushp and Lantao Liu
\thanks{Authors are with the Luddy School of Informatics, Computing, and Engineering, Indiana University, Bloomington, IN 47408 USA (e-mail: {\tt\small \{hjardali, mohamedi, dpushp, lantao\}@iu.edu}). \\
}}%
\definecolor{applegreen}{rgb}{0.8, 1, 0.0}
\definecolor{LightCyan}{rgb}{0.88,1,1}
\definecolor{atomictangerine}{rgb}{1.0, 0.6, 0.4}
\definecolor{amber}{rgb}{1.0, 0.75, 0.0}
\definecolor{aqua}{rgb}{0.0, 1.0, 1.0}
\definecolor{almond}{rgb}{0.94, 0.87, 0.8}
\definecolor{aquamarine}{rgb}{0.5, 1.0, 0.83}
\definecolor{babyblue}{rgb}{0.54, 0.81, 0.94}
\definecolor{babyblueeyes}{rgb}{0.63, 0.79, 0.95}
\definecolor{asparagus}{rgb}{0.53, 0.66, 0.42}
\definecolor{auburn}{rgb}{0.43, 0.21, 0.1}
\definecolor{brilliantlavender}{rgb}{0.96, 0.73, 1.0}
\definecolor{bittersweet}{rgb}{1.0, 0.44, 0.37}
\definecolor{blue-violet}{rgb}{0.54, 0.17, 0.89}
\definecolor{capri}{rgb}{0.0, 0.75, 1.0}
\definecolor{celadon}{rgb}{0.67, 0.88, 0.69}
\definecolor{darkcyan}{rgb}{0.0, 0.55, 0.55}
\definecolor{deepskyblue}{rgb}{0.0, 0.75, 1.0}
\definecolor{dogwoodrose}{rgb}{0.84, 0.09, 0.41}

\maketitle

\global\csname @topnum\endcsname 0
\global\csname @botnum\endcsname 0


\thispagestyle{empty}
\pagestyle{empty}

\begin{abstract}
Autonomous racing has attracted significant attention recently, presenting challenges in selecting an optimal controller that operates within the onboard system’s computational limits and meets operational constraints such as limited track time and high costs. This paper introduces a 
Linear Parameter-Varying Model Predictive Controller (LPV-MPC) for lateral control. Implemented on an IAC AV-24, the controller achieved stable performance at speeds exceeding \SI{160}{mph} (\SI{71.5}{\metre\per\second}).  We detail the controller design, the methodology for extracting model parameters, and key system-level and implementation considerations. Additionally, we report results from our final 
race run, providing a comprehensive analysis of both vehicle dynamics and controller performance. Moreover, a Python implementation of the framework can be accessed here:
~\url{https://tinyurl.com/LPV-MPC-acados}


\end{abstract}

\begin{IEEEkeywords}
Automated vehicles, autonomous racing, path tracking, MPC, lateral control, vehicle dynamics.
\end{IEEEkeywords}
\section{Introduction}\label{Introduction}
\IEEEPARstart{C}{ontrolling} 
an autonomous vehicle safely and efficiently across diverse driving conditions requires addressing key challenges in vehicle dynamics, actuation limits, and environmental uncertainties. A 
control strategy must account for the intricate coupling between longitudinal and lateral dynamics, nonlinear tire behavior, and variations in tire-road friction and actuator response. These challenges become even more pronounced at high speeds, where modeling errors can lead to higher tracking errors.  
Autonomous racing serves as an ideal testbed for evaluating control strategies
for vehicles operating at the limits of their physical capabilities. Racing competitions such as the Indy Autonomous Challenge (IAC) \cite{indyautonomouschallengeIndyAutonomous}
and the Abu Dhabi Autonomous Racing League (A2RL) \cite{a2rlDhabiAutonomous}
enable direct benchmarking of control approaches, providing valuable insights for real-world autonomous driving. 

In these competitions, teams participate in events such as time trials, passing, and multi-vehicle races, each requiring control strategies that balance precision, robustness, and adaptability. Our team, the IU-Luddy Autonomous Racing Team (IU-LART), competed in the Indy Autonomous Challenge (IAC) at the Indianapolis Motor Speedway (IMS) on September 6, 2024, and at the Las Vegas Motor Speedway (LVMS) on January 9, 2025~\cite{jardali2025zero}. Despite limited practice track time, we achieved a maximum speed of 163 mph at LVMS using an IAC AV-24 (see Fig.~1). In this article, we focus on the motion control algorithm deployed during the final IAC race at LVMS and analyze its performance under high-speed racing conditions.

\begin{figure}
    \centering
    \includegraphics[width=1\linewidth]{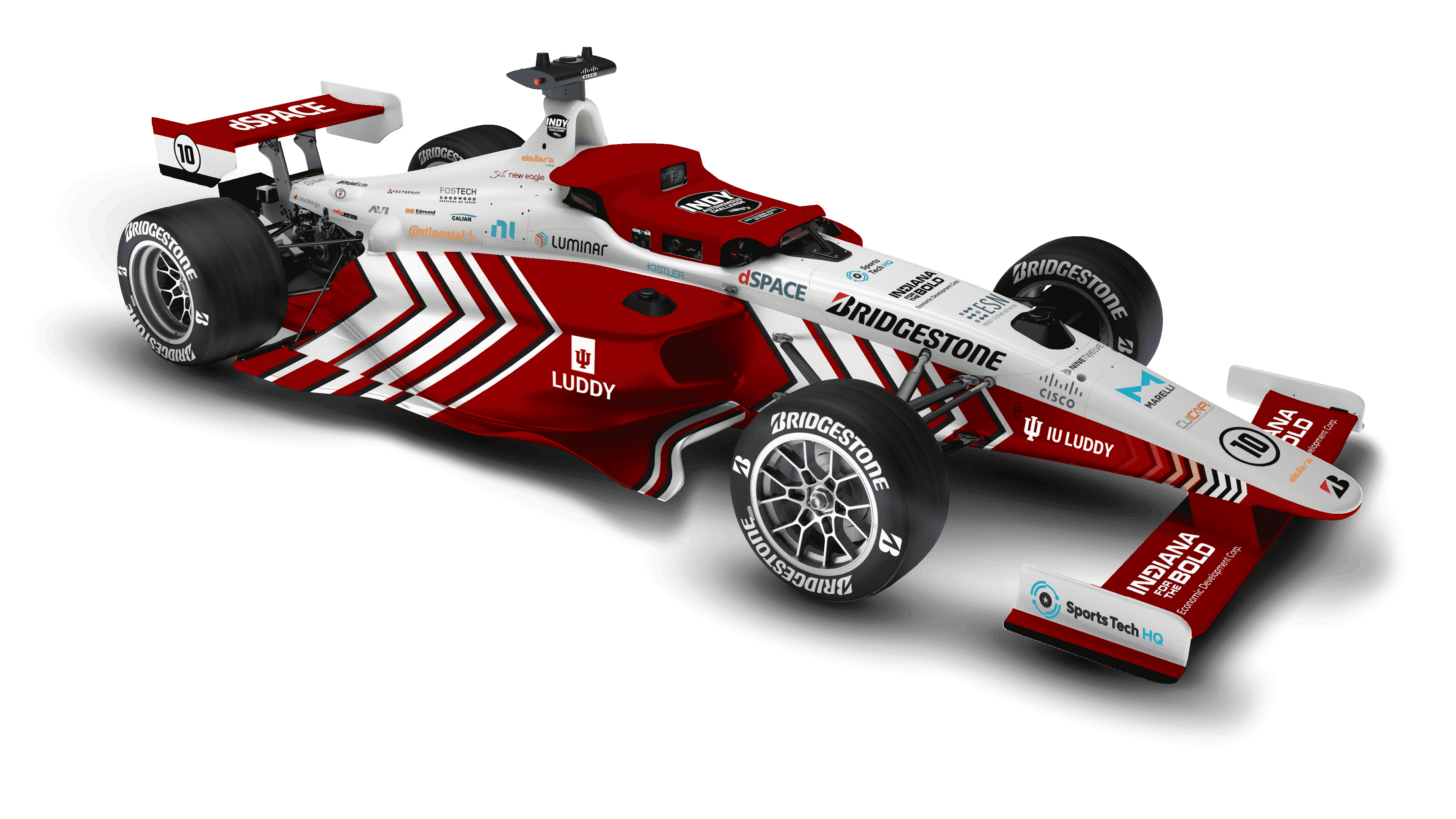}
    \caption{IU-LART IAC AV-24 autonomous racing vehicle \cite{jardali2025zero}.}
    \label{fig:IU_Luddy_av24-label}
\end{figure}

\section{Related Work}

Although oval-shaped racetracks provide a structured environment similar to freeways, high-speed driving introduces complexities that challenge conventional control strategies. At these speeds, nonlinear vehicle dynamics, model uncertainties, such as empirically estimated tire parameters, and external disturbances, including wind and track inclination, significantly impact stability. 
 
Consequently, it is essential to develop the right control strategy to achieve reliable path tracking and safe handling of the car in high-performance autonomous racing \cite{wischnewski2022tube, reda2024path, betz2022autonomous}.

Typically, the control module consists of two components: longitudinal control, which controls the car's longitudinal motion 
, and lateral control, which controls its steering. Consequently, control frameworks for path trajectory tracking are generally categorized as \emph{coupled} or \emph{decoupled} approaches \cite{paden2016survey, betz2022autonomous}.
Coupled strategies jointly optimize both lateral (steering) and longitudinal (acceleration/braking) inputs, capturing the intricate dependencies between vehicle dynamics to enhance tracking accuracy and stability \cite{ kabzan2019learning, alcala2020autonomous, raji2022motion, raji2023tricycle, wischnewski2022tube}. 
For instance, a learning-based MPC framework integrates Gaussian Process Regression (GPR) to take residual model uncertainty into account, allowing for improved lap time while maintaining the vehicle's safety \cite{kabzan2019learning}. 
In \cite{alcala2020autonomous}, the authors used an LPV-MPC for longitudinal and lateral control to track an offline-generated racing path; however, its applicability to high-speed racing environments remains uncertain, as it has not been validated on a full-scale autonomous racing car \cite{alcala2020autonomous}. 
Additionally, a tricycle model-based MPC explicitly accounts for the locked differential, improving stability and precision in high-speed maneuvers, especially on road-course tracks by coupling steering and acceleration dynamics \cite{raji2023tricycle}. However, these approaches introduce trade-offs, as the learning-based method incurs high computational costs and requires continuous data updates, and the tricycle model approach introduces modeling complexity and demands precise system identification.

Due to the challenges associated with coupled strategies, decoupled control remains widely adopted in both autonomous driving and racing scenarios, as it simplifies controller synthesis.
In decoupled control frameworks, the system dynamics are typically split into longitudinal and lateral motion, with longitudinal behavior often modeled as a linear first-order system, while lateral dynamics require more sophisticated control strategies due to their inherent complexity \cite{stano2023model, artunedo2024lateral}. This complexity necessitates greater attention in racing applications, where precise steering control is crucial for maintaining stability and achieving optimal trajectory tracking at high speeds.
Geometric-based algorithms for lateral motion, such as Pure Pursuit \cite{coulter1992implementation} and Stanley \cite{hoffmann2007autonomous}, are among the widely utilized path-tracking strategies due to their simplicity and computational efficiency. However, the performance of these kinematic models is limited by their core assumption of a no-slip condition. This premise, which assumes the tires roll perfectly without sliding, breaks down in high-speed racing where significant tire slip is essential for generating the large forces needed to maneuver at the limit of adhesion.

To overcome these limitations, optimization-based control strategies move beyond kinematic assumptions by using dynamic models that explicitly consider the relationship between tire slip and vehicle forces \cite{saba2024fast, chung2024autonomous}. In full-scale autonomous racing literature, \cite{saba2024fast} developed an LQR-based lateral tracking algorithm using a single-track lateral dynamics model and using a look-ahead search strategy to determine the target point along the path. Similarly, \cite{chung2024autonomous} employs an LQR controller but enhances it by incorporating the desired yaw rate and explicitly accounting for the track’s banking angle, improving trajectory tracking and stability.
Since both methods assume a constant longitudinal velocity, they incorporate a velocity-dependent weight scheduling. Additionally, due to the LQR framework, neither explicitly manages constraints.
MPC offers a more robust alternative to LQR for lateral vehicle control by addressing key limitations, including the assumption of constant longitudinal velocity and the lack of explicit constraint handling. 
Outside the full-scale autonomous racing literature, \cite{kebbati2021optimized} proposes an adaptive MPC with Laguerre functions for path tracking using an augmented single-track lateral dynamics model and optimization of the controller parameters and cost function weights with a Particle Swarm Optimization (PSO) algorithm.

In contrast to existing solutions, and given the constraints of limited physical access to the race car due to competition policies and restricted practice runs on the track, 
we present in this work a linear parameter-varying model predictive control (LPV-MPC) framework designed
for full-scale high-speed autonomous racing while highlighting the implementation steps. 
The contributions of this work can be summarized as:
\begin{enumerate}
    \item 
    We present an LPV-MPC framework for lateral control and path tracking,  based on a vehicle dynamic model that is formulated as an LPV system scheduled by longitudinal speed, trajectory curvature, and road banking.
    \item We conduct a real-world evaluation on a full-scale autonomous race car,
demonstrating stable control performance at speeds exceeding \SI{71.5}{\metre\per\second} (160 mph), coupled with a system identification detailed technique, and an in-depth analysis.
    This analysis includes a quantitative evaluation of performance and a discussion of the method's limitations.
    \item We release an open-source, Python-based implementation\footnote{
    ~\url{https://tinyurl.com/LPV-MPC-acados}}, providing a reference framework for the research community to advance autonomous racing research.
\end{enumerate}

The paper is organized as follows: Section \ref{Vehicle Model} presents the single-track lateral dynamic model and identification of parameters. Section \ref{MPC} details the proposed LPV-MPC formulation. Section \ref{experimental-validation} covers the experimental setup, followed by performance analysis. Finally, Section \ref{conclusion} summarizes key findings and outlines future research directions.
\section{Lateral Vehicle Dynamics}\label{Vehicle Model}
This section derives the single-track lateral dynamic model and extends it by incorporating the steering angle as a state variable. It also details the system identification method used to estimate the dynamic parameters.
\begin{figure}[!h]
\vspace{-5pt}
    \centering
    \includegraphics[scale=1]{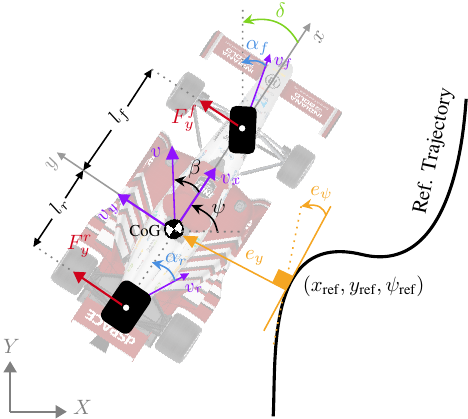}
    \vspace*{1pt}
    \caption{Overview of the single-track bicycle model, highlighting relevant forces, angles, and lateral errors in their positive directions.}
    \label{fig:dynamic_curv_frame-label}
    \vspace{-3pt}
\end{figure}

\subsection{Single-Track Lateral Dynamic Model}\label{Lateral Single-Track Dynamic Model}
Modeling vehicle dynamics accurately is challenging due to (i) varying road-tire friction and vertical loads; (ii) speed- and steering-dependent dynamics; and (iii) coupled lateral and longitudinal behavior. 
Owing to the typical symmetry of road vehicles, kinematic and dynamic bicycle (single-track) models are widely used to simplify vehicle dynamics, modeling the car as a rigid body with mass $m$ and yaw inertia $I_z$, as depicted in Fig.~\ref{fig:dynamic_curv_frame-label}.
The single-track model assumes perfect lateral symmetry; however, although the race car considered in this study has an asymmetrical weight distribution due to its oval-track setup, this effect is neglected in the proposed model.
To derive the governing equations of lateral dynamics, Newton's laws of motion are applied along the y- and z-axes, summing the lateral forces and yaw moment at the vehicle's center of gravity (CoG). The governing equations are given by
\begin{equation}
\label{eq:force_moment_balance}
\left\{
\begin{aligned}
    \sum F_y &= F^f_{y} + F^r_{y} = m\bigl(\ddot{y} + \dot{\psi} v_x\bigr), \\
    \sum M_z &= l_f F^f_{y} - l_r F^r_{y} = I_z \ddot{\psi},
\end{aligned}
\right.
\end{equation}
where \( F^f_{y} \) and \( F^r_{y} \) are the lateral forces at the front and rear tires, respectively,  
\( M_z \) is the resultant yaw moment,  
\( v_x \) is the longitudinal velocity,  
\(\dot{\psi}\) and \(\ddot{\psi}\) denote the yaw rate and yaw acceleration,  
\( l_f \) and \( l_r \) represent the distances from the CoG to the front and rear axles,  
and \( \ddot{y} \) is the lateral acceleration of the CoG.
To derive such a linear model, several fundamental assumptions are made. The slip angles are considered small, the tires operate within the linear region of the slip-angle–lateral-force relationship, and the road–tire friction coefficient
is assumed to be constant. 
Under these conditions, the lateral tire forces are approximated as
\begin{equation}
\begin{aligned}
F^f_{y} \approx 2 C_{f} \alpha_f, \quad    
F^r_{y} \approx 2 C_{r} \alpha_r,
\end{aligned}
\label{linear_forces_eq}
\end{equation}
where \( C_{f} \) and \( C_{r} \) represent the cornering stiffnesses of the front and rear tires, respectively.
To further describe the lateral dynamics, the front and rear slip angles, \( \alpha_f \) and \( \alpha_r \), are introduced. These angles characterize the deviation between the tire direction and the actual velocity vector at each axle (see Fig.~\ref{fig:dynamic_curv_frame-label}). 
Specifically, they are defined as $\alpha_f=\delta-\theta_f$ and $\alpha_r=-\theta_r$, where $\delta$ is the steering angle. Under small slip angles assumption, the front and rear velocity angles are approximated as $\theta_f \approx \beta + \tfrac{l_f\dot{\psi}}{v_x}$ and $\theta_r \approx \beta - \tfrac{l_r\dot{\psi}}{v_x}$, where $\beta=\arctan\!\big(\tfrac{v_y}{v_x}\big)\approx \tfrac{v_y}{v_x}$ is the sideslip angle, representing the lateral-to-longitudinal velocity ratio.

Building on this formulation, we define the model in terms of reference path tracking errors, as this approach minimizes complexity in the cost function, reduces computational overhead, and simplifies both the design and tuning of the controller. This choice is particularly important in high-speed racing, where precise tracking is crucial to maintaining stability and performance, as even small deviations can have significant effects.
The lateral position error \( e_y \) is defined as the signed perpendicular distance from the vehicle's CoG to the desired raceline, commonly referred to as the cross-track error (CTE).
Similarly, the heading error \( e_{\psi} \) is the difference between the vehicle’s yaw angle and the desired yaw angle, given by \( e_{\psi} = \psi - \psi_{\text{ref}} \), where \(\psi\) and \(\psi_{\text{ref}}\) are the global yaw angle of the car and the desired yaw angle at the nearest point on the raceline, respectively. 
To incorporate the desired yaw rate, we define \( \dot{\psi}_{\text{ref}} = v_x \kappa \), where \( \kappa \) is the instantaneous curvature of the local path.

Building on these tracking error definitions, the lateral vehicle dynamics in the curvilinear frame can be formulated as a state-space model in tracking error variables, leading to a linearized single-track model in matrix form given by
\begin{equation}
    \dot{\mathbf{x}} = A \mathbf{x} + B \mathbf{u} + C \dot{\psi}_{\text{ref}},
    \label{state_space_rep}
\end{equation}
where 
\( \mathbf{x} =
\begin{bmatrix}
e_y , \dot{e}_y, e_{\psi} , \dot{e}_{\psi}
\end{bmatrix}^{\top} \in \mathbb{R}^{4},  \) is the state vector representing the system's errors, \( \mathbf{u} = \delta \in \mathbb{R} \) is the control input, \( \dot{\psi}_{\text{ref}} \) is the external input derived from the reference trajectory, and \( A, B, C \) are matrices defined as
\[
A = 
\begin{bmatrix}
0 & 1 & 0 & 0 \\
0 & a_{22} & a_{23} & a_{24} \\
0 & 0 & 0 & 1 \\
0 & a_{42} & a_{43} & a_{44}
\end{bmatrix} \! \!, \! \! \! \! \! \!
\quad
B = 
\begin{bmatrix}
0 \\
b_{21} \\
0 \\
b_{41}
\end{bmatrix} \!, \! \!  \! \! \! \!
\quad
C =
\begin{bmatrix}
0 \\
a_{24} - v_x \\
0 \\
a_{44}
\end{bmatrix},
\]
with
\[
a_{22} = -\tfrac{2(C_{f}+C_{r})}{m\,v_x}\!, \! \!  \! \! \! \!
\quad
a_{23} = -v_x a_{22}, \! \!  \! \! \! 
\quad
a_{24} = \tfrac{-2C_{f}l_f + 2C_{r}l_r}{m\,v_x}, 
\]
\[
a_{42} = -\tfrac{2(C_{f}l_f - C_{r}l_r)}{I_z\,v_x},
\quad
a_{43} = -v_x a_{42}, \! \!  \! \! \!\!
\]
\[
a_{44} = -\tfrac{2(C_{f}l_f^2 + C_{r}l_r^2)}{I_z\,v_x},
\quad
b_{21} = \tfrac{2C_{f}}{m}, 
\quad
b_{41} = \tfrac{2C_{f}l_f}{I_z}.
\]

Controlling a high-speed vehicle on an oval track requires smooth and precise steering to maintain stability. To better capture actuator dynamics while systematically handling steering constraints, we extend the model presented in \eqref{state_space_rep} by incorporating the steering angle \( \delta \) as an additional state, while treating the steering rate \( \dot{\delta} \) as the control input, expressed as \( \mathbf{u} = [\dot{\delta}] \in \mathbb{R} \). This augmentation enables the explicit integration of steering constraints within the MPC framework, 
including lag and rate limitations.
As a result, the augmented model is formulated as:
\[
\tilde{A} = 
\begin{bmatrix}
A & B \\
0 & 0
\end{bmatrix}, 
\quad \! \! \! \!
\tilde{B} =
\begin{bmatrix}
0 , 0 , 0 , 0 , 1
\end{bmatrix}^{\top},
\quad \! \! \! \!
\tilde{C} = 
\begin{bmatrix}
C \\
0
\end{bmatrix}.
\]
Beyond steering dynamics, external factors such as road banking significantly influence lateral stability, particularly at high speeds. To account for these effects, we introduce the gravitational term \(g \sin(\phi)\) into the lateral dynamics, where \(\phi\) denotes the banking angle, reaching up to \(20^\circ\) at the testing track.
This additional lateral gravitational component redistributes the cornering forces, reducing the load borne solely by the tires and enabling higher sustained cornering speeds. However, this effect also necessitates explicit compensation in the control model to prevent 
oversteer or instability. 
By incorporating \( \phi \), the model accounts for the gravitational assist in turns, leading to a more accurate prediction of the required lateral tire forces.
Thus, the augmented state-space representation, including road banking effects, is formulated as: 
\begin{equation} 
\dot{\tilde{\mathbf{x}}} = \tilde{A} \tilde{\mathbf{x}} + \tilde{B} \mathbf{u} + \tilde{C} \dot{\psi}_{\text{ref}} + D_{\phi}, 
\label{aug_state_space_rep} 
\end{equation}
where the augmented state vector is defined as \( \tilde{\mathbf{x}} = [e_y, \dot{e}_y, e_{\psi}, \dot{e}_{\psi}, \delta]^{\top} \in \mathbb{R}^{5} \),  
and the disturbance matrix due to road banking is given by \( D_{\phi} = [0, g \sin(\phi), 0, 0, 0]^{\top} \in \mathbb{R}^{5} 
 \). Since \( D_{\phi} \) represents the effect of road banking, which is determined by track geometry rather than a controllable input, it is treated as a disturbance in the state-space model.  
 A detailed derivation of the model and a discussion of its assumptions are provided in Chapter 2 of \cite{rajamani2011vehicle}.

\subsection{Dynamics Parameters Identification}\label{Dynamics Parameters Identification} 
To effectively implement the model given in \eqref{aug_state_space_rep} in a real-world racing scenario, key model parameters must be estimated, particularly under high-speed conditions where precise vehicle dynamics modeling is essential for stability and performance. Among these parameters, the cornering stiffness values \( C_{f} \) and \( C_{r} \) play a crucial role, as they directly influence the lateral force estimation and controller response. However, obtaining accurate estimates of these parameters is challenging due to measurement noise, costly track operations, and an increased risk of crashes during parameter tuning.
In the autonomous racing literature, optimization-based methods \cite{becker2023model} and advanced software solutions \cite{raji2024er} have been utilized. Alternatively, learning-based approaches—such as hyperparameter optimization \cite{seong2023model} and physics-informed neural networks \cite{chrosniak2024deep}—offer data-driven strategies that improve model accuracy and reduce manual tuning.
These techniques typically rely on the nonlinear Magic Formula (Pacejka) tire model \cite{pacejka1992magic}, which is commonly formulated as:
\begin{equation}
\begin{split}
    F^i_y &= D^i_p \sin\Bigl[ C^i_p \arctan\bigl( B^i_p \alpha_i \\
    &\quad - E^i_p (B^i_p \alpha_i - \arctan(B^i_p \alpha_i))\bigr)\Bigr],
\end{split}
\label{pacejka_eq}
\end{equation}
where \(F^i_y\) is the lateral tire force and \(\alpha_i\) is the slip angle. The index \( i \) denotes the front (\( f \)) and rear (\( r \)) tires. The parameters are \(B^i_p\), the stiffness factor that scales slip; \(C^i_p\), the shape factor that influences curvature; \(D^i_p\), the peak factor that defines the maximum force; and \(E^i_p\), the curvature factor that adjusts peak behavior.

\begin{figure}[h] \vspace{-5pt}
    \centering
    \includegraphics[scale=0.75]{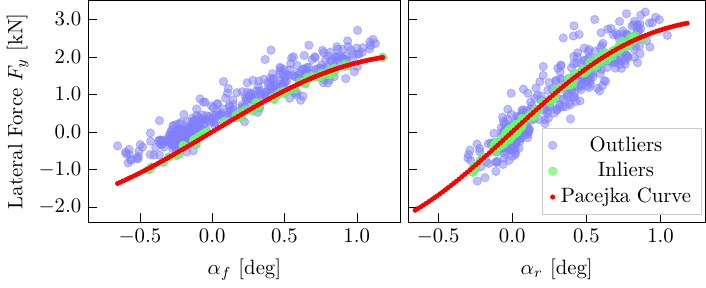}
    \caption{Measured cornering forces versus slip angles with the corresponding fitted Pacejka curve, obtained from the final IMS run at a longitudinal velocity of \SI{56}{\metre\per\second}. Note that the vertical offset in the Pacejka model (typically denoted as \(s_{hy}\)) was not accounted for.}
    \label{fig:IMS-dynamics-label}
  
\end{figure}

We determined the linear cornering stiffness values \(C_{f}\) and \(C_{r}\) following the paradigm of \cite{becker2023model}. In our work, Pacejka tire model parameters are estimated via an iterative expectation-maximization (EM) approach that employs a nonlinear least-squares optimization. The procedure iteratively refines the fit by excluding outliers identified using a Median Absolute Deviation (MAD) derived threshold until convergence is reached. The linear dynamics, used in \eqref{linear_forces_eq}, are then approximated using the relation \( C_{\text{linear}} = B_p C_p D_p \). To estimate our initial dynamics parameters, we used data from our previous highest-speed lap, in which the vehicle, controlled via Pure Pursuit, completed a lap at \SI{56}{\meter\per\second}.
The lateral forces at the front and rear tires were then computed from the measured lateral acceleration \(a_{y,\text{imu}}\) recorded by an Inertial Measurement Unit (IMU) using the relationships:
\begin{equation}
\begin{aligned}
\!
    F^f_{y} &= \frac{m l_r}{(l_r + l_f) \cos \delta} a_{y,\text{imu}}, \!\!\! \quad
    F^r_{y} &= \frac{m l_f}{l_r + l_f} a_{y,\text{imu}.}
\end{aligned}
\label{fyf_fyr_measured}
\end{equation}
Using \(\eqref{pacejka_eq}\) and \(\eqref{fyf_fyr_measured}\), we ran our optimization algorithm to determine the Pacejka parameters. These parameters were used to find the linear dynamics and served as initial inputs for tuning the MPC. 
Table \ref{IMS_pacejka_table} presents the optimized Pacejka parameters alongside the linear approximation. Fig.\,\ref{fig:IMS-dynamics-label} illustrates the fitted Pacejka curve overlaid on the measured data, showing the lateral forces versus slip angles for both the front and rear tires.
\begin{table}[h]
\vspace{7pt}
\caption{Initial Pacejka Parameters values using IMS data. $D_p$ is in Newtons (N), and $C_{\text{linear}}$ is in N/rad.}
    \centering
    \setlength{\tabcolsep}{4pt} 
    \renewcommand{\arraystretch}{1.1} 
    \begin{tabular}{l||cccc|c}
        \toprule
        \textit{Tires} & $B_p$ & $C_p$ & $D_p$ & $E_p$ & $C_{\text{linear}}$ \\
        \hline
         \hline
        \textit{Front }& 34.59 & 1.81 & 2100 & -1.0 & \SI{132}{k}\\
        \textit{Rear}  & 35.04 & 1.96 & 3036 & -0.26 & \SI{209}{k}\\
        \bottomrule
    \end{tabular}
    \label{IMS_pacejka_table}
\end{table}
As an additional point of comparison, we implemented a Physics-Informed Neural Network (PINN) approach following \cite{chrosniak2024deep}. Among the tested architectures, the PINN employing Gated Recurrent Units (GRUs) produced parameter estimates most consistent with those obtained from our EM optimization method. This agreement provides complementary evidence of methodological consistency across independent estimation frameworks. Subsequently, during practice sessions, these parameters were further fine-tuned. In Section~\ref{Results}, we present the estimated parameters derived from the data at \SI{162}{mph} using our EM optimization method.
\section{Path Tracking with LPV-MPC}\label{MPC}

Building on the previously defined augmented lateral dynamics model, we now formulate the LPV-MPC strategy for high-speed autonomous racing.
\subsection{Path Tracking Problem Formulation}
Consider a smooth reference trajectory consisting of \( N_r \) discrete reference points, denoted as the reference state \( \mathbf{x}_{\text{ref}, k} = \begin{bmatrix} x_{\text{ref}, k}, y_{\text{ref}, k}, \psi_{\text{ref}, k}, v_{\text{ref}, k} \end{bmatrix}^\top \), which specifies a desired position \( (x_{\text{ref}, k}, y_{\text{ref}, k}) \) in the global frame, yaw angle \( \psi_{\text{ref}, k} \), and velocity \( v_{\text{ref}, k} \) along the planned trajectory.  
Let us assume that the current vehicle state is given by \( \mathbf{x}_k = \begin{bmatrix} x_k, y_k, \psi_k, v_{x,k}, v_{y,k}, \dot{\psi}_k \end{bmatrix}^\top \), where \( x_k \) and \( y_k \) represent the vehicle's position in the global frame, \( \psi_k \) is the yaw angle, \( v_{x,k} \) and \( v_{y,k} \) are the longitudinal and lateral velocities, respectively, and \( \dot{\psi}_k \) is the yaw rate.   
The reference trajectory is generated by the planning module, while the current vehicle state is provided by the state estimation module, as described in Section~\ref{System}. 
Since the system operates in the error space, the state vector \( \tilde{\mathbf{x}}_k \), which is formally defined in Section~\ref{Vehicle Model}, is derived from the lateral components of the actual vehicle state \( \mathbf{x}_k \) and the reference state \( \mathbf{x}_{\text{ref}, k} \).  
Considering a finite time-horizon \( N \), the sequence of control inputs is defined as \( \mathbf{U} = \left[\mathbf{u}_{0}, \mathbf{u}_{1}, \dots, \mathbf{u}_{N-1}\right]^{\top} \in \mathbb{R}^{N} \), while the corresponding state trajectory of the system is \( \tilde{\mathbf{X}} = \left[\tilde{\mathbf{x}}_{0}, \tilde{\mathbf{x}}_{1}, \dots, \tilde{\mathbf{x}}_{N}\right]^{\top} \in \mathbb{R}^{5(N+1)} \). 

The objective of the lateral controller is to minimize the tracking errors in \(\tilde{\mathbf{x}}_k\)
, ensuring accurate path following within the given system constraints, while accounting for variations in curvature, speed, and external disturbances to maintain stability.
This is achieved by generating an optimal sequence of steering inputs \( \mathbf{U} \) over a finite prediction horizon \( N \).  
To formulate the optimization problem, the continuous-time LPV model \eqref{aug_state_space_rep} is first discretized. 
Assuming the control inputs and scheduling parameters are held constant over a sampling interval $T_s$
(a Zero-Order Hold assumption), the exact discretization of \eqref{aug_state_space_rep} yields the following discrete-time LPV model:
\begin{equation}
\mathbf{\tilde{x}}_{k+1} = \tilde{A}_d(\mathbf{p}_k)\,\mathbf{\tilde{x}}_k + \tilde{B}_d(\mathbf{p}_k)\,\mathbf{u}_k + \mathbf{\tilde{E}}_d(\mathbf{p}_k)
\label{dis-dyn-model}
\end{equation}
where the discrete-time matrices $\tilde{A}_d$, $\tilde{B}_d$,
along with the affine term $\mathbf{\tilde{E}_d}$, which accounts for the effects of $\dot{\psi}_{\text{ref}}$ and $\phi$, 
are derived from their continuous-time counterparts and depend on the scheduling parameters vector \( \mathbf{p} = [v_x , \kappa , \phi]^\top \in \mathbb{R}^3 \), which consists of the longitudinal velocity \( v_x \), road curvature \( \kappa \), and road banking angle \( \phi \). 
This parameter-dependent structure allows the proposed control strategy to be inherently adaptive 
to varying speed profiles and track geometry. Consequently, the matrices $\tilde{A}_d$, $\tilde{B}_d$ and $\mathbf{\tilde{E}_d}$
are updated within the optimization problem at each time step $k$.
The scheduling parameter vector $\mathbf{p}$ is required for the entire prediction horizon. At the current time step ($k = 0$), $\mathbf{p}$ is known exactly from real-time sensor measurements. For future steps within the horizon ($k = 1, \dots, N$), a prediction strategy is employed: the future longitudinal velocity $v_x$ and road curvature $\kappa$ are extracted from the pre-computed reference trajectory, while the road banking angle $\phi$ is assumed to remain constant at its most recently measured value.

As a result, this formulation yields an LPV-MPC path-tracking algorithm that conditions its optimal control problem (OCP) on the real-time measurement of the scheduling parameters, overcoming limitations imposed by fixed model assumptions, such as assuming constant velocity or neglecting banking and race-line curvature effects, that were assumed previously in full-scale autonomous racing (see \cite{saba2024fast, chung2024autonomous}).

\subsection{Constrained Control Problem}

To formulate the path-tracking task, we define a constrained nonlinear least-squares optimal control problem. The objective function incorporates penalties on state tracking errors, control effort, and the side-slip angle:
\begin{equation}
J = \sum_{k=0}^{N-1} 
\left(
\| \tilde{\mathbf{x}}_k \|_Q^2 
+ \| \mathbf{u}_k \|_R^2 
+ Q_{\beta} \, \beta_k(\tilde{\mathbf{x}}_k)^2
\right),
\label{cost_j}
\end{equation}
where the weighting matrix $Q \in \mathbb{R}^{5 \times 5}$ is diagonal and penalizes deviations in the state variables, while $R \in \mathbb{R}$ is a positive scalar that penalizes control effort. 

Following insights from~\cite{raji2023tricycle}, we include a side-slip penalty to enhance vehicle stability by discouraging large slip angles. The side-slip angle is computed from the predicted states and is defined as
$
\beta_k(\tilde{\mathbf{x}}_k) 
= \arctan\!\left( \frac{\dot e_{y,k}}{v_{x,k}} \right),
$
where $\dot e_{y,k}$ is the predicted lateral velocity error and $v_{x,k}$ is the longitudinal velocity treated as a scheduling parameter. Because $\beta_k$ depends nonlinearly on the decision variable $\tilde{\mathbf{x}}_k$, its inclusion renders the overall objective function nonlinear.
The optimal control problem can finally be formulated as:
\begin{subequations} \label{eq:lpv_mpc_problem}
\begin{align}
    \min_{\tilde{\mathbf{X}}, \mathbf{U}} \quad & J(\mathbf{\tilde{x}}, \mathbf{u}) \label{eq:lpv_mpc_obj} \\
    \text{s.t.} \quad 
    & \mathbf{\tilde{x}}_{k+1} = \tilde{A}_d(\mathbf{p}_k)\,\mathbf{\tilde{x}}_k 
      + \tilde{B}_d(\mathbf{p}_k)\,\mathbf{u}_k + \nonumber  \\
    &  \quad\quad\quad\;\;   \mathbf{\tilde{E}}_d(\mathbf{p}_k), 
     \quad  \quad k = 0, \dots, N-1, 
    \label{eq:lpv_mpc_dynamics} \\
    & \delta_{\min} \leq \delta_k \leq \delta_{\max}, 
      \enspace \quad k = 0, \dots, N-1, 
    \label{eq:lpv_mpc_delta_bounds} \\
    & \mathbf{u}_{\min} \leq \mathbf{u}_k \leq \mathbf{u}_{\max},
      \quad k = 0, \dots, N-1.
    \label{eq:lpv_mpc_delta_rate}
\end{align}
\end{subequations}

Our proposed LPV-MPC strategy solves this optimization problem by minimizing the objective function $J$ \eqref{eq:lpv_mpc_obj}, taking into account the system dynamics with time-varying parameters $\mathbf{p}$ \eqref{eq:lpv_mpc_dynamics}, the steering constraints \eqref{eq:lpv_mpc_delta_bounds}, which are implicitly incorporated within \(\tilde{\mathbf{x}}\), and the control input limits \eqref{eq:lpv_mpc_delta_rate}, which impose rate constraints on \( \mathbf{u}_k = \dot{\delta}_k \). This process generates an optimal control sequence \( \mathbf{U}\), from which only the first control input \( \mathbf{u}_0 \) is applied to the autonomous racing vehicle. 
It is worth noting that the terminal cost in \eqref{cost_j} is omitted since our empirical evaluations confirmed that its inclusion yielded negligible performance benefits for the chosen prediction horizon, while omitting it improved computational efficiency.
\section{Experimental Validation}\label{experimental-validation}
In this section, we validate our proposed LPV-MPC strategy through real-world experiments. We first outline the hardware and software architecture, followed by the experimental results from our 
final run at the racing competition, along with a detailed analysis and discussion of the system's performance.

\subsection{Experimental Setup}\label{System}
To validate our proposed control strategy, we conducted experiments using an IAC AV-24, a full-scale autonomous racing vehicle (see Fig.~\ref{fig:IU_Luddy_av24-label}). The vehicle is equipped with advanced hardware and a 
Drive-by-Wire (DBW) system for autonomous operation, including a dSPACE Autera AutoBox powered by an Intel Xeon 3GHz 12-core CPU and an NVIDIA A5000 GPU. 
It runs Ubuntu 22.04 with CycloneDDS for real-time communication and leverages ROS2 as the core development framework.
For completeness, we briefly outline state estimation and planning, as they are essential for running our controller effectively, although these modules are beyond the scope of this work. The state estimation module fuses GNSS, IMU, wheel odometry, and steering angle data using an Unscented Kalman Filter (UKF) \cite{yu2023fast}, thereby achieving accurate real-time localization. The planning module consists of both offline and online components: the offline planner generates a raceline using the minimum curvature method \cite{heilmeier2020minimum} as a global path, while the online planner continuously tracks the nearest path segment and publishes a local-frame trajectory. 
\begin{figure}[!t]
\vspace{-2pt}
    \centering
    \includegraphics[scale=0.9]{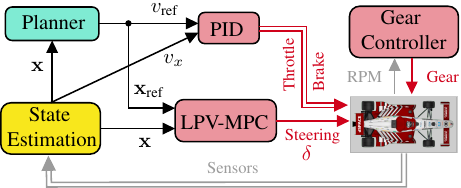}
    \vspace*{1pt}
    \caption{High-level software architecture of our autonomous racing system for a single-car format \cite{jardali2025zero}.}
    \label{fig:software_architecture}
\end{figure}

As illustrated in Fig.~\ref{fig:software_architecture}, the vehicle's control architecture follows a decoupled design, comprising separate longitudinal and lateral control modules, with well-defined signal flow between components.
These modules process the vehicle's primary control inputs, which include throttle ($0$–$100$\%), brake pressure (in kPa), gear command (ranging from gear $1$ to $6$), and steering angle (in degrees). 
The longitudinal controller regulates the vehicle's speed by modulating throttle and brake inputs, while also executing gear shifts.
In this work, a Proportional-Integral-Derivative (PID) controller is employed to track the desired velocity by modulating throttle and brake inputs, while gear shifting is regulated based on engine RPM.
The lateral control module is implemented using our proposed LPV-MPC, designed to handle the vehicle's lateral dynamics in high-speed scenarios by generating the desired steering rate \(\dot{\delta}\), which is then integrated to produce the steering angle $\delta$ required to track the reference trajectory provided by the planning module.
For safety, a Pure Pursuit (PP) controller continuously runs as a backup in case the optimization fails due to constraint violations, numerical instability at low speeds, or exceeding real-time computation limits.

The control system operates in real-time, with the planning and control loop running at \SI{50}{\hertz}, while state estimation updates at \SI{125}{\hertz} providing precise localization.
The optimal control problem in~\eqref{eq:lpv_mpc_problem} is solved using \textit{acados}~\cite{verschueren2022acados}, with code adapted from~\cite{kloeser2020nmpc}. Because the side-slip term $\beta_k$ depends nonlinearly on the predicted states, the formulation constitutes a constrained nonlinear least-squares optimal control problem. We therefore employ the real-time iteration sequential quadratic programming (SQP–RTI) scheme, in which a single quadratic programming (QP) subproblem is constructed and solved at each control step using HPIPM~\cite{frison2020hpipm}. A Gauss–Newton approximation is used for the Hessian of the nonlinear least-squares objective. 
The problem is discretized over a prediction horizon of $T = 1.6$\,s with $N = 45$ intervals and solved using a fixed-step ERK integrator ($\epsilon = 10^{-4}$).

Due to velocity-dependent terms in the model (Section~\ref{Lateral Single-Track Dynamic Model}), numerical ill-conditioning arises as $v_x \to 0$. The controller therefore switches to a backup Pure Pursuit controller below \SI{45}{mph} (\SI{20}{\meter\per\second}). The weighting matrices $Q$, $R$, and $Q_\beta$ were tuned in simulation and further refined offline between track sessions alongside the vehicle’s dynamic parameters.

\subsection{Results}\label{Results}
In this section, we present results from our on-track practice sessions and the final racing competition, in which our team participated exclusively in the time-trials format. Leading up to the event, we completed 16 on-track sessions, each averaging about 25 minutes. 
Despite the considerable challenges of high logistical costs, limited track access, and expenses related to car repairs, we maximized every practice session to focus on parameter tuning, race-line validation, and other competition-specific preparations.
It is important to note that, given the competitive nature of autonomous racing, teams do not share code or internal implementations. As a result, direct modular or quantitative comparisons with other teams’ approaches are not possible. Instead, in this subsection, we highlight the results from our final racing competition, which represent the most compelling outcome of our efforts—our fully autonomous run on an oval track during the challenge. 

During the final competition, the run included 12 high-speed laps. On the final lap, we achieved our highest recorded speed of \SI{163.1}{mph} (\SI{72}{\meter\per\second}), with a maximum absolute cross-track error of \SI{1.6}{\meter}. The best lap time was \SI{34.18}{\second} with an average speed of \SI{157.98}{mph} (\SI{70.6}{\meter\per\second}). It is worth noting that the maximum velocity reached during the practice runs was only \SI{145}{mph} (\SI{64.8}{\metre\per\second}).   

\subsubsection{Tracking Errors:}
\begin{figure}
    \centering
    \begin{subfigure}{0.82\linewidth}
        \centering
        \hspace*{-23pt}\includegraphics[scale=0.82,trim=0cm 0.2cm 0cm 0cm, clip]{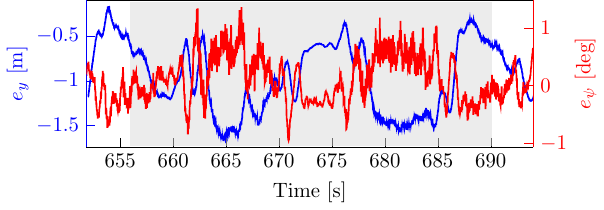}
        \caption{Cross-track error $e_y$ and heading error $e_\psi$.}
        \vspace*{4pt}
        \label{fig:errors_vs_time-label}
    \end{subfigure}
    \vspace*{3pt}
    
    \begin{subfigure}{0.82\linewidth}
        \centering
        
        \hspace*{-42pt}\includegraphics[scale=0.82,trim=0cm 0.2cm 0cm 0cm, clip]{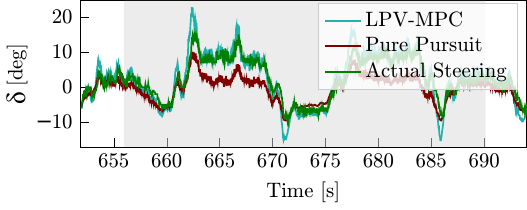}
        \caption{MPC, Pure-Pursuit \textit{(backup)}, and applied steering angles.}
        \vspace*{1pt}
        \label{fig:mpc_vs_pp_steering-label}
    \end{subfigure}
    \vspace*{3pt}

    \begin{subfigure}{0.82\linewidth}
        \centering
        \hspace*{-14pt}\includegraphics[scale=0.82,trim=0cm 0.2cm 0cm 0cm, clip]{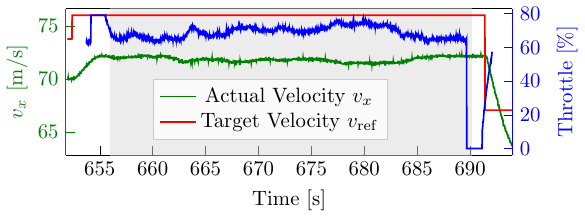}
        \caption{Target velocity, actual velocity, and applied throttle.}
        \label{fig:velocity_throttle_vs_time-label}
        \vspace*{4pt}
        
    \end{subfigure}
    \caption{Experimental results for path tracking during our highest speed lap. The highlighted areas mark the start and finish of the lap.}
    \label{fig:tracking-errors-figures}
\end{figure}
The tracking errors consist of the cross-track error \(e_y\) measured in \(\si{\metre}\) and the heading error \(e_\psi\) measured in \(\si{\deg}\).
Additionally, we present the commanded and actual longitudinal velocities, measured in \(\si{\metre\per\second}\). Fig.~\ref{fig:errors_vs_time-label} shows the cross-track and heading errors during the high-speed lap. The mean cross-track error is approximately \(-1.02\)\,\si{\metre}, with a standard deviation of \(0.38\)\,\si{\metre}. The heading error remains within \(\pm 1^\circ\) throughout the lap. 
As shown in Fig.~\ref{fig:errors_vs_time-label}, the cross-track error $e_y$ remains negative over the entire lap—indicating the vehicle consistently tracked to the left of the reference line—and this persistent behavior
is attributable to either a steering offset in the hardware or the race car’s non-symmetrical weight configuration for oval racing (or both), neither of which is accounted for in our model.
In Fig.~\ref{fig:mpc_vs_pp_steering-label}, we compare the steering angle commands from the LPV-MPC controller and the backup PP controller with the actual steering applied by the vehicle. The LPV-MPC commands range from \(-20^\circ\) to \(20^\circ\), indicating no steering constraint violations. However, at $t=663, 671, 678,$ and \SI{686}{\second} there is a large discrepancy between the applied steering and the LPV-MPC command, likely due to the latter’s rapid rate of change. This suggests the steering rate-limit constraints require further tuning.
Notably, the MPC exhibits higher responsiveness, resulting in larger steering commands, whereas the PP controller takes a more gradual approach, demonstrating the MPC's aggressive approach to minimizing lateral deviations.
Finally, Fig.~\ref{fig:velocity_throttle_vs_time-label} shows the target velocity, actual velocity, and applied throttle. A persistent velocity error of approximately \SI{4}{\metre\per\second} is observed, indicating that the longitudinal controller requires adjustment to improve speed tracking.
\begin{figure}
    \centering
        \hspace*{-0pt}\includegraphics[scale=0.82]{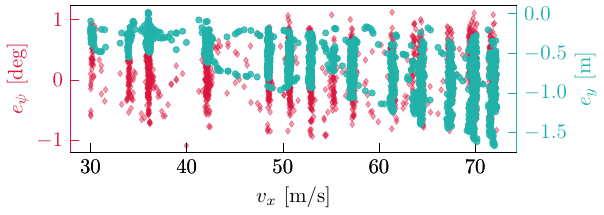}
    \caption{Tracking errors vs. velocity during the complete run.}
    \label{fig:errors_vs_velocity-label}
\end{figure}

Fig.~\ref{fig:errors_vs_velocity-label} illustrates the relationship between tracking errors and longitudinal velocity $v_x$ across the 12 high-speed laps of the competition run with our proposed control strategy. The heading error \(e_\psi\) shows no clear dependence on velocity and remains within a relatively narrow band, whereas the cross-track error \(e_y\) increases approximately linearly with velocity. 
By characterizing $e_y$ as a function of longitudinal velocity $v_x$, we anticipated tracking error at race speeds and tailored our strategy.
\subsubsection{Dynamics and Model Errors:}
\begin{figure}[!t]
    \centering
    \hspace{-4pt}\includegraphics[scale=0.75]{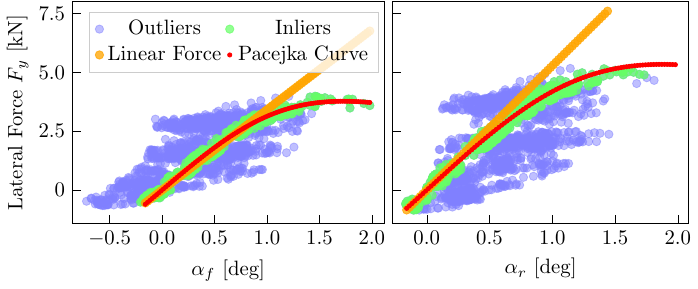}
    \caption{Measured cornering forces $F_y$ versus slip angles, $\alpha_f$ and $\alpha_r$, with the corresponding fitted Pacejka curve, obtained from the final 
    run at a longitudinal velocity of \SI{72}{\metre\per\second}. The \textit{Linear Force} curve depicts the estimated force computed using the cornering stiffness parameters that were determined during the practice runs and employed during the final run.}
    \label{fig:dynamics-label}
\end{figure}
\begin{table}[h]
\caption{Pacejka parameter values obtained via EM optimization using our final run 
data. $D_p$ is in Newtons [N], and $C_{\text{linear}}$ is in [N/rad]. 
Note that these parameters have not been tested with the proposed controller.
}
    \centering
    \setlength{\tabcolsep}{4pt} 
    \renewcommand{\arraystretch}{1.1} 
    \begin{tabular}{l||cccc|c}
        \toprule
       \textit{Tires} & $B_p$ & $C_p$ & $D_p$ & $E_p$ & $C_{\text{linear}}$ \\
        \hline \hline
        \textit{Front} & 22.30 & 2.00 & 3885.85 & -1.00 & \SI{173}{k} \\
        \textit{Rear}  & 26.08 & 2.00 & 5342.89 & -1.00 & \SI{278}{k} \\
        \bottomrule
    \end{tabular}
    \label{VEGAS_pacejka_table}
\end{table}

During the final run, the vehicle achieved speeds exceeding those attained in practice. As shown in Fig.~\ref{fig:dynamics-label}, slip angles reached up to \SI{2}{\deg}, briefly entering the nonlinear region. The figure shows that, at higher slip angles, the linear force model (orange; calibrated on practice data) overestimates the lateral forces estimated from the on-board IMU.
Additionally, Table~\ref{VEGAS_pacejka_table} lists the Pacejka parameters identified from the same IMU-derived final-run data using the method in Section~\ref{Dynamics Parameters Identification}.

\begin{figure}
    \centering
    \includegraphics[scale=0.82]{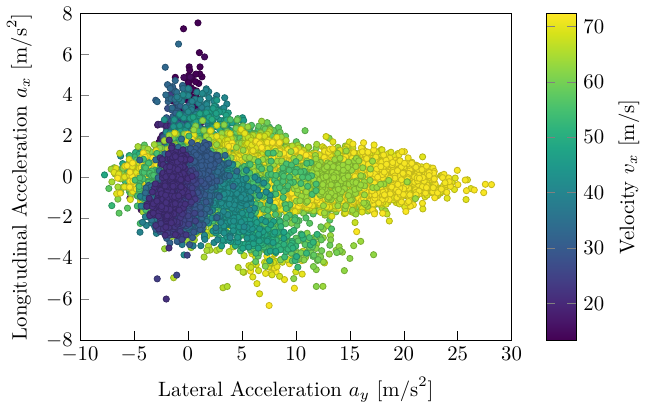}
    \caption{Lateral vs. longitudinal acceleration (g-g diagram) during the complete run.  Accelerations were measured using the on-board IMU.}
    \label{fig:gg_diagram-label}
\end{figure}
\begin{figure}
    \centering
    \includegraphics[scale=.82]{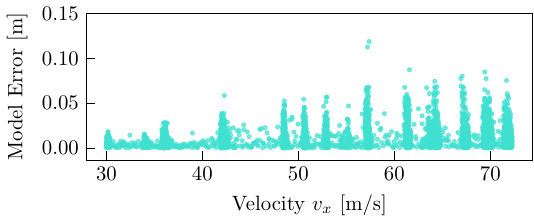}
    \caption{Lateral model error vs. velocity during the final run.}
    \label{fig:model_error_vs_velocity-label}
\end{figure}

In Fig.~\ref{fig:gg_diagram-label}, the vehicle’s lateral and longitudinal accelerations (g-g diagram) over the entire run are shown, colored by longitudinal velocity $v_x$. 
At the highest longitudinal speeds, the car achieved a lateral acceleration $a_y$ of up to \SI{27}{\metre\per\second\squared} (yellow points, $v_x \approx \SI{70}{\metre\per\second}$)
It is worth noting that we adopted a conservative raceline with large boundary margins to ensure safety, resulting in higher curvature. Despite these constraints,
the controller maintained stable performance throughout.

Fig.~\ref{fig:model_error_vs_velocity-label} presents the lateral model error, computed offline as the one-step-ahead difference between the MPC-predicted cross-track error $\hat e_y$ and the actual $e_y$ from the state-estimation output, thus, the lateral model error can be formulated as $e_y^{\mathrm{mdl}}=\hat e_y-e_y$.
Beyond approximately \SI{55}{\meter\per\second}, the maximum model error stabilizes around \SI{0.06}{\meter}. 
However, during high-acceleration events (e.g., at \SI{42}{\meter\per\second} and \SI{57}{\meter\per\second}), the error spikes, reaching a peak of \SI{0.14}{\meter}. 
This occurs when aggressive throttle inputs induce longitudinal slip in the rear tires, reducing their lateral force capacity—a dynamic coupling not fully captured by our model.

\subsubsection{Computation Time:}

\begin{figure}[H]
    \vspace{-0.05in}
    \centering
    \includegraphics[scale=0.82]{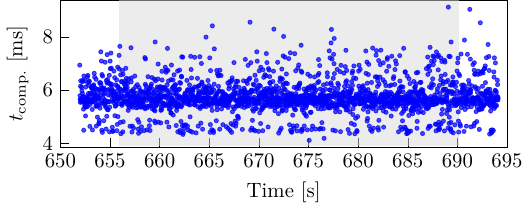}
    \caption{Computation time $t_\text{comp.}$ of the control loop during the high-speed lap.}  
    \label{fig:computation_time-label}
\end{figure}
Fig.~\ref{fig:computation_time-label} demonstrates the real-time performance of the controller, with a mean computation time of \SI{5.78}{\milli\second}. Some spikes are attributed to other processes such as data logging, which intermittently increase CPU contention. Despite these transient delays, the control loop consistently operates within a \SI{10}{\milli\second} cycle time. These low computation times confirm the feasibility of operating at 100~Hz, and future iterations of this work will target this frequency to enhance command smoothness and reactivity.

\subsection{Discussion}\label{Discussion}
Despite the encouraging results, several limitations of this work warrant mention:
\subsubsection{Cross-Track Error:} The observed cross-track error was moderate to high. The primary causes were an unmodeled steering offset in the hardware, which we cannot calibrate or access; the car’s asymmetric lateral weight distribution; inaccuracies in extracting vehicle-dynamic parameters; and suboptimal tuning of the weighting matrices. Due to limited track time and competition requirements, we prioritized controller stability over minimizing tracking error. While this trade-off was acceptable for single-car racing, it may be unsuitable in multi-vehicle contexts requiring precise trajectory tracking. With more time, improved parameter identification and better tuning of the weighting matrices and constraints could have reduced the cross-track error.

\subsubsection{Terminal Cost:} 

Although removing the terminal cost had no noticeable effect in our tests, retaining it is theoretically advisable. Future work should include a robust terminal cost to improve closed-loop stability.

\subsubsection{Longitudinal Control:} 
For single-car oval racing with minimal speed variation, our simple velocity-based controller was adequate. However, for road-course tracks or multi-vehicle racing, a more advanced longitudinal control approach is required.

\subsubsection{Limitations of the Linear Model:}

The LPV formulation treats $v_x$, $\kappa$, and $\phi$ as exogenous scheduling parameters, neglecting cross-sensitivities and preventing the controller from exploiting their variation within the prediction horizon. While track banking kept operation largely within the tire model’s linear region, a flat track would induce nonlinear behavior at lower speeds. Future work should consider a fully nonlinear MPC formulation to better capture these coupled effects.

\section{Conclusion}\label{conclusion}
In this study, we introduced the Linear Parameter-Varying Model Predictive Controller (LPV-MPC) for lateral control, as implemented by our team.
The proposed controller enabled the racing vehicle to maintain bounded tracking errors at speeds exceeding \SI{160}{mph} (\SI{71.5}{\metre\per\second}). Furthermore, we detailed a methodology for extracting dynamic parameters, supported by a comprehensive analysis of both vehicle dynamics and controller performance, while also addressing the limitations inherent in our approach. To promote reproducibility and stimulate further research, the Python-based source code for the framework is openly available. Future research efforts will focus on mitigating the identified limitations to further enhance performance and broaden the framework’s applicability to increasingly complex autonomous racing environments, such as multi-vehicle racing on road-course tracks. In addition, we will conduct a systematic sensitivity analysis to rigorously evaluate the controller’s robustness to model uncertainties and varying environmental conditions.

\section{Acknowledgments}
We thank all the students and researchers who contributed to the development of the software stack, especially Youwei Yu and Mahmoud Ali for their work on key modules and architectures. We are also grateful to the IAC organization for supporting this work. ChatGPT was used to help improve the presentation of the manuscript.

\bibliographystyle{IEEEtran}
\bibliography{references}            

\end{document}